\documentclass[letterpaper]{article} 
\usepackage{aaai23}  
\usepackage{times}  
\usepackage{helvet}  
\usepackage{courier}  
\usepackage[hyphens]{url}  
\usepackage{graphicx} 
\usepackage{natbib}  
\usepackage{caption} 
\usepackage{algorithm}
\usepackage{algorithmic}
\usepackage{amsthm}
\usepackage{amsmath}
\usepackage{amssymb}

\usepackage{booktabs}
\usepackage{multirow}
\usepackage{diagbox}
\usepackage{pifont}

\usepackage{newfloat}
\usepackage{listings}
\DeclareCaptionStyle{ruled}{labelfont=normalfont,labelsep=colon,strut=off} 
\floatstyle{ruled}
\newfloat{listing}{tb}{lst}{}
\floatname{listing}{Listing}
\title{NQE: N-ary Query Embedding for Complex Query Answering over Hyper-Relational Knowledge Graphs}
\author{
    Haoran Luo\textsuperscript{\rm 1},
    Haihong E\textsuperscript{\rm 1}\thanks{Corresponding author.},
    Yuhao Yang\textsuperscript{\rm 2},
    Gengxian Zhou\textsuperscript{\rm 1},\\
    Yikai Guo\textsuperscript{\rm 3},
    Tianyu Yao\textsuperscript{\rm 1},
    Zichen Tang\textsuperscript{\rm 1},
    Xueyuan Lin\textsuperscript{\rm 1},
    Kaiyang Wan\textsuperscript{\rm 1}
}
\affiliations{
    \textsuperscript{\rm 1}School of Computer Science, Beijing University of Posts and Telecommunications, Beijing, China\\
    \textsuperscript{\rm 2}School of Automation Science and Electrical Engineering, Beihang University, Beijing, China\\
    \textsuperscript{\rm 3}Beijing Institute of Computer Technology and Application, Beijing, China\\
    \{luohaoran, ehaihong\}@bupt.edu.cn, yangyuhao818@buaa.edu.cn

}

\usepackage{bibentry}

\begin{document}

\maketitle

\begin{abstract}
Complex query answering (CQA) is an essential task for multi-hop and logical reasoning on knowledge graphs (KGs). Currently, most approaches are limited to queries among binary relational facts and pay less attention to n-ary facts (n$\geq$2) containing more than two entities, which are more prevalent in the real world. Moreover, previous CQA methods can only make predictions for a few given types of queries and cannot be flexibly extended to more complex logical queries, which significantly limits their applications. To overcome these challenges, in this work, we propose a novel \textbf{N}-ary \textbf{Q}uery \textbf{E}mbedding (\textbf{NQE}) model for CQA over hyper-relational knowledge graphs (HKGs), which include massive n-ary facts. The NQE utilizes a dual-heterogeneous Transformer encoder and fuzzy logic theory to satisfy all n-ary FOL queries, including existential quantifiers ($\exists$), conjunction ($\wedge$), disjunction ($\vee$), and negation ($\neg$). We also propose a parallel processing algorithm that can train or predict arbitrary n-ary FOL queries in a single batch, regardless of the kind of each query, with good flexibility and extensibility. In addition, we generate a new CQA dataset WD50K-NFOL, including diverse n-ary FOL queries over WD50K. Experimental results on WD50K-NFOL and other standard CQA datasets show that NQE is the state-of-the-art CQA method over HKGs with good generalization capability. Our code and dataset are publicly available.
\end{abstract}

\section{Introduction}

\begin{figure}[t]
\centering
\includegraphics[width=8.5cm]{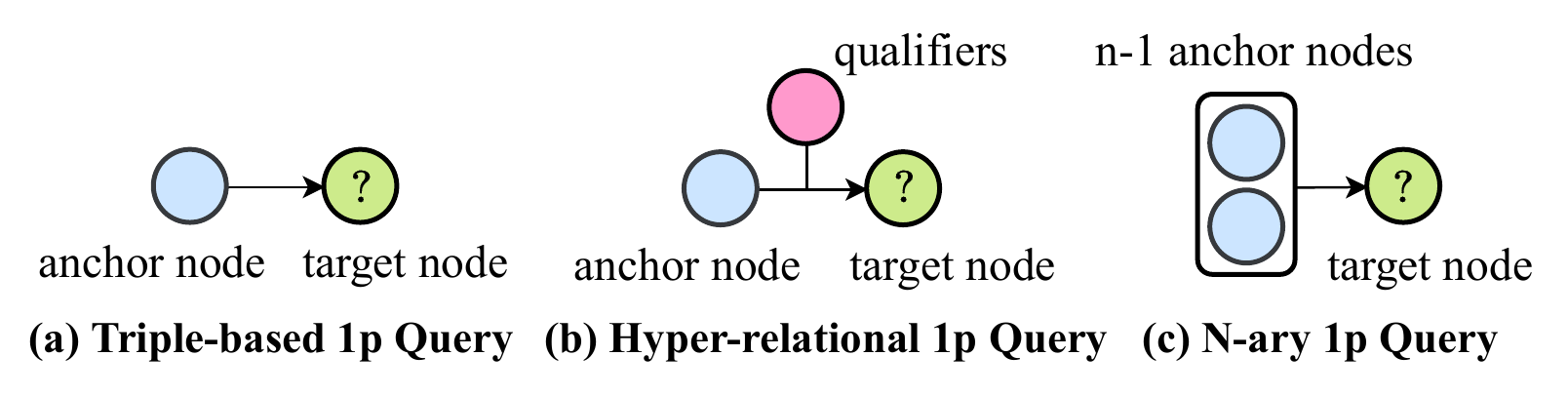}
\caption{The difference between (a) triple-based 1p query, (b) hyper-relational 1p query, and (c) n-ary 1p query, three one-hop (1p) queries.}
\label{f2}
\end{figure}

Complex query answering (CQA) is essential for multi-hop and logical reasoning over Knowledge Graphs (KGs). Traditional CQA methods obtain answers through graph database query languages such as SPARQL, for which challenges remain, such as high time complexity and missing edges. To tackle these challenges, some approaches based on query embedding (QE)~\citep{GQE,Q2B,BetaE,CQD,ConE,FLEX,FuzzQE} propose to embed queries and entities into the same latent space according to computational graphs, obtaining query answers by computing similarity scores. However, most QE approaches are limited to queries among binary relational facts in the form of triples (\textit{subject}, \textit{relation}, \textit{object}), as in Figure~\ref{f2}(a).

In modern large-scale KGs, e.g., Wikidata~\citep{Wikidata} and Freebase~\citep{Freebase}, in addition to binary relational facts, n-ary facts ($n\geq2$) are also abundant, which have a novel hyper-relational representation of one primary triple plus n-2 attribute-value qualifiers $((s,r,o),\{(a_i:v_i)\}^{n-2}_{i=1})$~\citep{HINGE}. For instance, a real-world fact, \textit{``Micheal Jackson received Grammy Award for Song of the Year for work of We are the World."}, can be described as a 3-ary fact ((\textit{Micheal Jackson, award received, Grammy Award for Song of the Year}), \{( \textit{for work}: \textit{We are the World} )\}). Based on this representation, StarQE~\citep{StarQE} extends triple-based queries to hyper-relational queries by introducing qualifiers information into the relations of triples as in Figure~\ref{f2}(b), which is the only approach currently attempting to solve the CQA problem for hyper-relational knowledge graphs (HKGs).

\begin{figure*}[h!t]
\centering
\includegraphics[width=18.1cm]{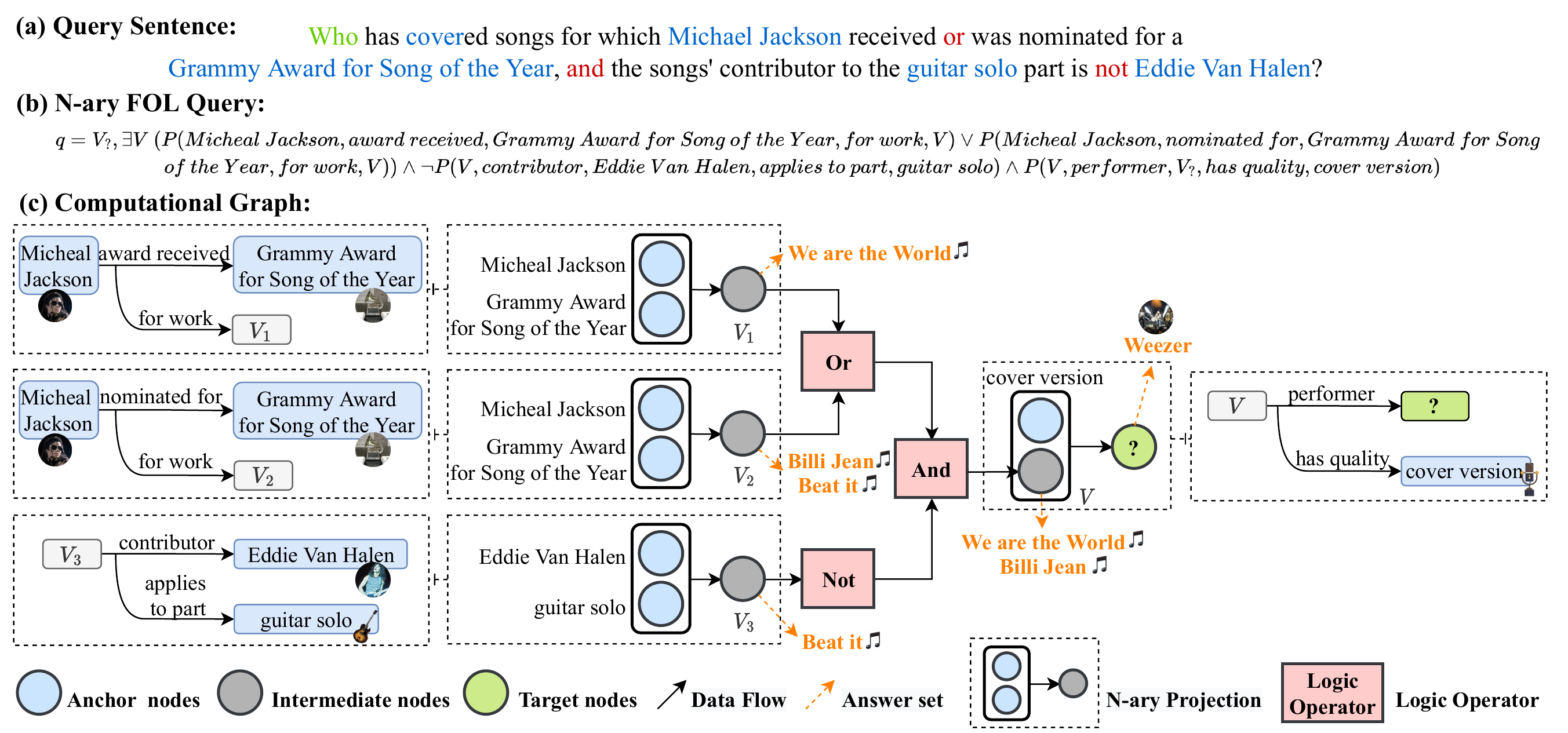}
\caption{An example of N-ary FOL query over HKG. (a) We found an example of a CQA query in Wikidata~\citep{Wikidata}, the largest hyper-relational knowledge base, and represented it by a query sentence. (b) Using the N-ary FOL query definition proposed in Sec.~\ref{s3}, we obtain the standard form of the query. (c) The computational graph corresponds to the N-ary FOL query and represents the topology of multi-hop and logical reasoning, which contains four atomic operations: n-ary projection, conjunction (And), disjunction (Or), and negation (Not).}
\label{f1}
\end{figure*}

Despite this, there are still three main challenges with CQA over HKGs. (1) \textbf{N-ary fact queries.} Previous methods could only introduce qualifiers into binary relational fact queries and query the tail entities, but not those entities in the n-ary facts, such as those in the qualifiers, which limits the scope of the n-ary fact query. (2) \textbf{Complex logical queries over HKGs.} Since HKGs have a unique structure, previous approaches can only handle existential quantifiers ($\exists$) and conjunction ($\wedge$) by directed acyclic graphs (DAGs) using message passing and fusion of nodes. Nevertheless, it still cannot define complex logical queries such as disjunction ($\vee$) and negation ($\neg$) in first-order logic (FOL). (3) \textbf {Diversity of query types.} There are many possibilities for complex logical queries of n-ary facts due to uncertainties in the number of elements, query locations, and logical issues. However, previous methods use a set of predefined query types, and different queries are handled separately, making CQA methods inflexible and difficult to apply practically.

To overcome these challenges, we propose a novel \textbf{N}-ary fact \textbf{Q}uery \textbf{E}mbedding (\textbf{NQE}) approach\footnote{\url{https://github.com/LHRLAB/NQE}} over HKGs, which aims to solve the CQA task for n-ary facts. First and foremost, we define N-ary FOL Query over HKGs, which allows an entity at any position ($s$, $o$, or $v_i$ in qualifier$_i$) in an n-ary fact to be queried by projection with other n-1 anchor entities, as in Figure~\ref{f2}(c). Due to the heterogeneity between entities and relations within n-ary facts in HKGs, inspired by GRAN~\citep{HAN,RPRT,GRAN}, NQE utilizes a dual-heterogeneous Transformer encoder to implement n-ary projection. Moreover, NQE initializes entity embedding and query embedding in the fuzzy space and defines three logical operators that conform to the logical axioms interpretably to implement all logical paradigms of n-ary FOL queries, as in Figure~\ref{f1}. Furthermore, to address the difficulty of diverse query types, NQE also employs a parallel processing algorithm that can train or predict arbitrary n-ary FOL queries in a single batch with good flexibility and scalability, regardless of each query.

In addition, we generate a new CQA dataset WD50K-NFOL, including diverse n-ary FOL queries over WD50K, and design various experiments for n-ary FOL queries over the new and other standard CQA datasets. The experimental results indicate that NQE can perform all FOL queries for n-ary facts, reaching a state-of-the-art level. We also set up experiments to verify that NQE still has good performance trained with only the one-hop n-ary entity prediction task, which reflects good generalization ability. 

Our main contributions are summarized as follows:

\begin{itemize}
\item We define the CQA task of N-ary FOL queries on HKGs for the first time.
\item We propose an N-ary QE model NQE capable of handling arbitrary entity queries in n-ary facts and satisfying all FOL queries for the first time. 
\item We propose a parallel processing algorithm that handles diverse queries in CQA tasks flexibly and extensively.
\item We generate a new HKG dataset with n-ary FOL queries, and the experimental results indicate that NQE is the state-of-the-art QE model on HKGs with good generalization capability.
\end{itemize}

\section{Related Work}
In this section, we present some studies closely related to our work, divided by n-ary fact representation learning and query embedding.
\subsection{N-ary Fact Representation Learning} 
Early knowledge graphs (KGs) studies were based on binary relational facts. Since binary relation oversimplifies the data properties and n-ary facts are abundant in real-world KGs, on the one hand, some approaches, m-TransH~\citep{m-TransH}, RAE~\citep{RAE}, n-TuckER~\citep{n-TuckER}, HypE~\citep{HypE}, and BoxE~\citep{BoxE}, consider n-ary facts of the form $r(e_1,e_2\ldots)$ and are extended directly from binary relational models~\citep{TransH, TuckER, SimplE}. On the other hand, some approaches, NaLP~\citep{NaLP}, RAM~\citep{RAM}, and NeuInfer~\citep{NeuInfer}, regard n-ary facts as role-value pairs of the form $r({r_1:e}_1,r_2:e_2\ldots)$, treating all entities equally and considering their role semantics additionally. 

To satisfy real-world fact structures, recent approaches, HINGE~\citep{HINGE}, StarE ~\citep{StarE}, GRAN~\citep{GRAN}, and Hyper2~\citep{Hyper2}, adapt a hyper-relational representation in the form of one primary triple plus several attribute-value qualifiers $r((e_1,r_1,e_2),\{r_2:e_3,r_3:e_4\ldots\})$ to solve the problem of one-hop link prediction on hyper-relational KGs (HKGs), which consist of n-ary facts with this representation. Moreover, StarQE~\citep{StarQE} uses StarE as the basic query graph encoder and a message-passing mechanism to extend the hyper-relational link prediction into multi-hop queries, but they still cannot handle queries for entities in qualifiers as variable nodes. 

For the first time, our model NQE implements a multi-hop logical query for entities in any position in n-ary facts as intermediate variable nodes using a dual-heterogeneous Transformer encoder.

\subsection{Query Embedding}

Query Embedding (QE) refers to a novel neural-symbolic method for complex query answering (CQA) problems, which can be divided into three types of logical queries in order of development. \textbf{(1)} GQE~\citep{GQE}, and BIQE~\citep{BIQE} are QE models for conjunctive queries, containing existential quantifiers ($\exists$) and conjunction($\wedge$). StarQE~\citep{StarQE} proposes hyper-relational conjunctive queries to introduce conjunctive queries into the hyper-relational knowledge graph. \textbf{(2)} Query2Box~\citep{Q2B}, EMQL~\citep{EMQL}, HypE~\citep{HypE}, CQD~\citep{CQD}, and PERM~\citep{PERM} handle arbitrary logical queries involving disjunction($\vee$) in addition to $\exists$ and $\wedge$, which are called existentially positive first order (EPFO) queries. \textbf{(3)} BetaE~\citep{BetaE}, ConE~\citep{ConE}, NewLook~\citep{NewLook}, Fuzz-QE~\citep{FuzzQE}, and GNN-QE~\citep{GNN-QE} can handle all first-order logic (FOL) with logical operations including $\exists$, $\wedge$, $\vee$, and negation($\neg$).

However, no QE model can handle the full range of FOL queries over HKGs for n-ary facts. Hence, in this paper, we define n-ary FOL queries and propose the NQE model attempting to solve the CQA problem for all n-ary FOL queries over HKGs.

\section{Preliminaries}
\label{s3}

In this section, we define three fundamental components of our work: the HKG and N-ary Fact, the N-ary Projection, and the N-ary FOL Query, followed by a problem definition.

\subsubsection{Definition 1: HKG and N-ary Fact.} 
A hyper-relational knowledge graph (HKG)
$\mathcal{G}=\{\mathcal{E},\mathcal{R},\mathcal{F}\}$ consists of an entity set $\mathcal{E}$, a relation set $\mathcal{R}$, and an n-ary fact (n$\geq$2) set $\mathcal{F}$. Each n-ary fact $f^n \in \mathcal{F}$ consists of entities $\in \mathcal{E}$ and relations $\in \mathcal{R}$, which is represented according to a hyper-relational paradigm~\citep{HINGE}:
\begin{equation}
\label{eq1}
f^n=\left\{
\begin{aligned}
&(s,r,o)  , & n=2, \\
&(s,r,o,\{a_i,v_i\}_{i=1}^{n-2})  , & n>2,
\end{aligned}
\right.
\end{equation}
where $(s,r,o)$ represents the main triple, $\{a_i,v_i\}^{n-2}_{i=1}$ represents n-2 auxiliary attribute-value qualifiers. Specifically, subject $s$, object $o$, and value $v_i$ are in the n-ary fact's entity set $E_f=\{e_p\}_{p=1}^n\subseteq\mathcal{E}$, main relation $r$ and attribute $a_i$ are in the n-ary fact's relation set $ R_f=\{r_q\}_{q=1}^{n-1}\subseteq\mathcal{R}$. More clearly, an n-ary fact can be understood as a sequence of n entities and (n-1) relations arranged alternately:
\begin{equation}
\begin{aligned}
f^n&(E_f,R_f)=\\&(e_1,r_1,e_2,r_2,e_3,\ldots,r_{n-1},e_{n}), n\geq2,
\end{aligned}
\label{eqfn}
\end{equation}
where $e_1=s,\ e_2=o,\ e_{p}=v_{p-2}\ (3\leq p\leq n) \in E_f$, $r_1=r,\ r_{q-1}=a_{q-2}\ (3\leq q\leq n) \in R_f$. We can also observe that binary relational facts are a special case of n-ary facts.

\subsubsection{Definition 2: N-ary Projection.}
N-ary projection denotes the process in which we lose one entity out of the n entities in an n-ary fact in HKGs and use the other (n-1) entities and (n-1) relations to predict these entities. We determine whether it is a subject, object, or a value of some qualifier by the position $p$ of the missing entity answer list $[e_p]$. 
\begin{equation}
[e_p]=P(f^n(E_f,R_f),?)=\left\{
\begin{aligned}
&P_s(f^n,?),&p=1,\\
&P_o(f^n,?),&p=2,\\
&P_{v_{p-2}}(f^n,?),&p\geq3,\\
\end{aligned}
\right.
\end{equation}
where $p=1,2,\ldots,n$ is the index of the missing entity in an n-ary fact $f^n$. For example, if a missing entity is a value $v_4$, the n-ary projection is represented as $[e_{6}]=P_{v_4}(e_1,r_1,e_2,r_2,e_3,\ldots,r_5,?,\ldots,r_{n-1},e_n)$. Specifically, the n-ary projection is equivalent to the one-hop n-ary entity prediction task on HKGs~\citep{GRAN}.

\subsubsection{Definition 3: N-ary FOL Query.} 
An n-ary FOL query $q$ consists of a non-variable anchor entity set $V_a \subseteq \mathcal{E}$, existentially quantified bound variables $V_1, V_2,\ldots, V_k$, target variable $V_?$, non-variable relation $R \subseteq \mathcal{R}$ and n-ary fact $f^n \in \mathcal{F}$ in the HKG, and logical operations of existential quantification ($\exists$), conjunction ($\wedge$), disjunction ($\vee$), and negation ($\neg$). The disjunctive normal form (DNF) of a query $q$ is defined as:
\begin{equation}
q[V_?]= V_?, \exists V_1,V_2,\ldots,V_k:c_1\vee c_2 \vee\ldots\vee c_n,
\end{equation}
\begin{equation}
c_i=d_{i1}\wedge d_{i2}\wedge\ldots\wedge d_{im},
\end{equation}
where $c_i$ represents conjunctions of one or more literals $d_{ij}$, and literal $d_{ij}$ represents an atomic formula or its negation:
\begin{equation}
d_{ij} = \left\{
\begin{aligned}
&P(f^n(E_f,R_f),V), \\
&\neg P(f^n(E_f,R_f),V),\\ 
&P(f^n(E_f\uplus V^{\prime}, R_f),V), \\
&\neg P(f^n(E_f\uplus V^{\prime}, R_f),V),
\end{aligned}
\right.
\end{equation}
where $E_f \subseteq V_a$ is the fact's anchor entity set, $R_f \subseteq R$ is the fact's non-variable relation set, $V \in \{V_?, V_1, V_2,\ldots, V_k\}$ represents the predicting entity in n-ary projection $P(f^n,V)$, $V^{\prime} \subseteq \{V_1, V_2,\ldots, V_k\}$ is a variable set to be combined ($\uplus$) with anchor entity set $E_f$ for n-ary projection, and $V \notin V^{\prime}$ makes sure there are no circular queries.

\subsubsection{Problem Definition.}
Given an incomplete HKG $\mathcal{\widehat{G}}=\{\mathcal{E},\mathcal{R},\mathcal{\widehat{F}}\}$ and an n-ary FOL Query $q$, the goal of CQA is to find a set of entities $S_q = \{a|a \in \mathcal{E}, q[a]\ is\ true\}$, where $q[a]$ is a logical formula that replaces the target variable $V_?$ of the query with the entity $a$.
A complex n-ary FOL query can be considered as a combination of multiple atomic subqueries $S_{q_i}$ in a computational graph, including n-ary $Projection$ operation and three logical operations, $Conjunction$, $Disjunction$, and $Negation$ which is defined as follows:
\begin{equation}
\begin{aligned}
&Projection(\{S_{q_i}\}_{i=1}^{n-1})&=&\ P(f^n(\uplus_{i=1}^{n-1} S_{q_i},R_f),?),\\
&Conjunction(\{S_{q_i}\}_{i=1}^m)&=&\ \cap_{i=1}^{m} S_{q_i},\\
&Disjunction(\{S_{q_i}\}_{i=1}^m)&=&\ \cup_{i=1}^{m} S_{q_i},\\
&Negation(S_q)&=&\ \ \mathcal{E}\backslash S_q.
\end{aligned}
\end{equation}

\section{Methodology}
In this section, we introduce our proposed model \textbf{N}-ary \textbf{Q}uery \textbf{E}mbedding (\textbf{N}QE) that can solve all n-ary FOL queries on HKGs.
\subsection{Embeddings with Fuzzy Vectors}
Fuzzy logic~\citep{logic1} is a logic that further extends the classical logic with only true or false to fuzzy sets with infinitely many values in the interval [0,1] to describe the degree of membership between elements and fuzzy sets. Fuzzy logic handles the fuzziness of many real-world objects with stronger logical representations and inherits the classical logic axioms. To satisfy all n-ary FOL queries, inspired by~\citep{FuzzQE}, we embed queries, entities, and relations in HKG as fuzzy logic space vectors $[0, 1]^d\subset R^d$ as $\boldsymbol{q}$, $\boldsymbol{x^e}$, and $\boldsymbol{x^r}$, respectively. As mentioned in Section~\ref{s3}, given a complex n-ary FOL query $\boldsymbol{q}$, 
NQE decomposes it into some intermediate subqueries $\boldsymbol{q}_1,\boldsymbol{q}_2,\ldots,\boldsymbol{q}_k$, and a final subquery $\boldsymbol{q}_{[tar]}$ by topological ordering of computational graph. The subqueries correspond to four basic operations: n-ary $Projection$, $Conjunction$, $Disjunction$, and $Negation$, which are defined as neural operations of the n-ary projector ($\mathcal{P}$), and three logical operators (conjunction $\mathcal{C}$, disjunction $\mathcal{D}$, negation $\mathcal{N}$), respectively, with fuzzy vector embeddings.

\subsection{N-ary Projection}

One of the most important atomic subqueries in processing N-ary FOL queries is the n-ary projection operator ($\mathcal{P}$), which derives the query embedding of missing positions from n-1 entity embeddings (or combined with variable query embeddings) and n-1 relation embeddings in an n-ary fact $f^n$. We replace the missing positions with [MASK], and the input is a sequence $\boldsymbol{X}=\big[\boldsymbol{x^e_s},\boldsymbol{x^r_r},\boldsymbol{x^e_o},\{\boldsymbol{x^r_{a_i}},\boldsymbol{x^e_{v_i}}\}_{i=1}^{n-2}\big]=\big[\boldsymbol{x}_1,\boldsymbol{x}_2,\ldots,\boldsymbol{x}_L\big] \in \mathbb{R}^{L\times d}$, while the output is the variable query embedding $\boldsymbol{q}_{[var]}$ as follows:
\begin{equation}
\begin{aligned}
\boldsymbol{q}_{[var]}=\mathcal{P}(\boldsymbol{X}).
\end{aligned}
\end{equation}

There are two kinds of elements in a sequence of an n-ary fact, containing entities and relations, showing the node heterogeneity. Also, the heterogeneous sequence has 14 kinds of edges between elements $s-r, s-o, r-o, s-a, s-v, r-a, r-v, o-a, o-v, a_i-a_j, o_i -o_j, a_i-o_i, a_i-o_j$, showing the edge heterogeneity. As we all know, Transformer~\citep{Transformer} shows great performance in sequence learning. However, traditional Transformer can represent node heterogeneity by absolute position encoding but loses the relative position information. Inspired by \citep{HAN}, \citep{RPRT}, and \citep{GRAN}, we design a novel dual-heterogeneous Transformer encoder to model both the absolute position information and the relative position information of the heterogeneous sequence with both node heterogeneity and edge heterogeneity:
\begin{equation}
m_{i j}=\frac{\left(\mathbf{W}_{role(i)}^{Q} \boldsymbol{x}_{i}+\mathbf{b}_{i j}^{Q}\right)^{\top}\left(\mathbf{W}_{role(j)}^{K} \boldsymbol{x}_{j}+\mathbf{b}_{i j}^{K}\right)}{\sqrt{d}},
\end{equation}
\begin{equation}
\alpha_{i j}=\frac{\exp \left(m_{i j}\right)}{\sum_{k=1}^{n} \exp \left(o_{i k}\right)},
\end{equation}
\begin{equation}
\tilde{\boldsymbol{x}}_{i}=\sum_{j=1}^{n} \alpha_{i j}\left(\mathbf{W}_{role(j)}^{V} \boldsymbol{x}_{j}+\mathbf{b}_{i j}^{V}\right),
\end{equation}
where $role(i)=\mathbf{r}$, if $2|i$, else $\mathbf{e}$. We first use ($\mathbf{W}_{\mathbf{e}}^{Q}, \mathbf{W}_{\mathbf{e}}^{K}, \mathbf{W}_{\mathbf{e}}^{V}) \in \mathbb{R}^{d \times d}$ and ($\mathbf{W}_{\mathbf{r}}^{Q}$, $\mathbf{W}_{\mathbf{r}}^{K}$, $\mathbf{W}_{\mathbf{r}}^{V})\in \mathbb{R}^{d \times d}$ to project embeddings of entities and relations into the same latent space, respectively, which reflects the node heterogeneity. Then we initialize 14 kinds of ($\mathbf{b}_{i j}^{Q},  \mathbf{b}_{i j}^{K}, \mathbf{b}_{i j}^{V}) \in \mathbb{R}^{d}$ for the edge heterogeneity, which is indexed by i-th and j-th elements. $m_{i j}$ is importance between one element in sequence $\boldsymbol{x}_i$ and another $\boldsymbol{x}_j$, $\alpha_{i j}$ is the semantic attention in the n-ary fact. Finally, the sequence embeddings are updated as $\tilde{\boldsymbol{x}_{i}} \in \mathbb{R}^{d}$ by one layer of the encoder. After the K-layer encoder, we take the updated vector at the [Mask] position $\tilde{\boldsymbol{x}}_{[Mask]}$ and pass it through the MLP, LayerNorm, and sigmoid layers as the output:
\begin{equation}
\mathcal{P}(\boldsymbol{X})=\sigma(LN(\mathbf{MLP}(\tilde{\boldsymbol{x}}_{[Mask]}))),
\end{equation}
where $\mathbf{MLP}:\mathbb{R}^{d}\rightarrow\mathbb{R}^{d}$, LN is a layernorm layer, and $\sigma$ is an activation function which guarantees that the result is also in the fuzzy logic space between [0,1].

\subsection{Logic Operators}

To handle the three logical atomic subqueries, we follow \citep{CQD}, \citep{FuzzQE}, and \citep{GNN-QE}, using fuzzy logic t-norm and t-conorm theory to introduce three common methods for logical operators: product logic, Gödel logic, and Lukasiewicz logic. Take product logic as an example; we have:
\begin{equation}
\begin{aligned}
&\mathcal{C}(\boldsymbol{q}_{a},\boldsymbol{q}_{b})&=&\ \boldsymbol{q}_{a}\circ\boldsymbol{q}_{b},\\
&\mathcal{D}(\boldsymbol{q}_{a},\boldsymbol{q}_{b})&=&\ \boldsymbol{q}_{a}+\boldsymbol{q}_{b}-\boldsymbol{q}_{a}\circ\boldsymbol{q}_{b},\\
&\mathcal{N}(\boldsymbol{q})&=&\ \boldsymbol{1}-\boldsymbol{q},
\end{aligned}
\end{equation}
where $\mathcal{C}$, $\mathcal{D}$, and $\mathcal{N}$ are three logical operators: conjunction, disjunction, and negation. $\boldsymbol{1}$ is an all-one fuzzy vector, while $\circ$, $+$, and $-$ are element-wise product, plus, and minus, respectively. Unlike GQE~\citep{GQE}, Q2B~\citep{Q2B}, BetaE~\citep{BetaE}, etc., these logical operators satisfy all the logical axioms, including t-norm logic axioms, t-conorm logic axioms, and negation logic axioms, to be more expressive and interpretable.

In the practical use of NQE, we further extend this formula to make $\mathcal{C}$, and $\mathcal{D}$ accept m subqueries (m$\geq$2) to obtain the output:
\begin{equation}
\boldsymbol{q}_{[var]}=\left\{
\begin{aligned}
&\mathcal{C}\Big(\{\boldsymbol{q}_{i}\}_{i=1}^m\Big)&=&\ \prod_{i=1}^m\boldsymbol{q}_{i},\\
&\mathcal{D}\Big(\{\boldsymbol{q}_{i}\}_{i=1}^m\Big)&=&\ \sum_{i=1}^m\boldsymbol{q}_{i}-\sum_{1 \leq i<j \leq m}\boldsymbol{q}_{i} \circ \boldsymbol{q}_{j}\\&&&+\sum_{1 \leq i<j<k \leq m}\boldsymbol{q}_{i} \circ \boldsymbol{q}_{j} \circ \boldsymbol{q}_{k}\\&&&-\cdots+(-1)^{m-1}\prod_{i=1}^m\boldsymbol{q}_{i},\\
&\mathcal{N}\Big(\boldsymbol{q}\Big)&=&\ \boldsymbol{1}-\boldsymbol{q},
\end{aligned}
\right.
\end{equation}
where $\prod,\sum$ represents the element-wise product and plus with multiple variable query embeddings.

\subsection{Model Learning}
\label{4.4}

With the above four atomic operators $\mathcal{P},\mathcal{C},\mathcal{D},\mathcal{N}$, we can derive the query embedding of intermediate and target nodes $\boldsymbol{q}_1,\boldsymbol{q}_2,\ldots,\boldsymbol{q}_k, \boldsymbol{q}_{[tar]}$ based on the topological sequence of the computational graph.

\subsubsection{Similarity Function.}
We define the similarity function $S$ to determine the similarity between the variable query embedding and the entity embedding as follows:
\begin{equation}
\begin{aligned}
S(\boldsymbol{q})=softmax(\boldsymbol{q}\mathbf{W}_e^\top),\\
\end{aligned}
\end{equation}
where $\mathbf{W}_e\in [0,1]^{|\mathcal{E}|\times d}$ shares the initial entity embedding matrix, $\boldsymbol{q}\mathbf{W}_e^\top\in \mathbb{R}^{|\mathcal{E}|}$ denotes the predicted value of similarity between $\boldsymbol{q}$ and the entity embedding, and $S(\boldsymbol{q})\in [0,1]^{|\mathcal{E}|}$ is obtained after softmax operation, which denotes the similarity probability of $\boldsymbol{q}$ for each entity in HKG. In this way, we can get the best answer to an arbitrary variable query $\boldsymbol{q}$ in $\{\boldsymbol{q}_1,\boldsymbol{q}_2,\ldots,\boldsymbol{q}_k, \boldsymbol{q}_{[tar]}\}$ by sorting the similarity probabilities $S(\boldsymbol{q})$. By setting a threshold, we can also derive a group of answers and probabilities with good interpretability.

\subsubsection{Loss Function.}
Finally, the similarity between the target variable query and all entities is $S(\boldsymbol{q}_{[tar]})$, using cross entropy to calculate the training loss with the labels:
\begin{equation}
\mathcal{L}=\sum_{t=1}^{|\mathcal{E}|}{\mathbf{y}_t\log S(\boldsymbol{q}_{[tar]})},
\end{equation}
where $\mathbf{y}_t$ is the $t$-th entry of the label $\mathbf{y}$. Since each query may have more than one answer, we follow~\citep{GRAN} and use label smoothing to define the label y. We set $\mathbf{y}_t = 1- \epsilon$ for the target entity and $\mathbf{y}_t=\frac{\epsilon}{|\mathcal{E}|-1}$ for each of the other entities, with label smoothing rate $\epsilon \in (0,1)$.

\subsubsection{Parallel Processing.}
To facilitate the parallel training and testing of different classes of N-ary FOL queries, we designed a parallel processing algorithm with one n-ary projection operator $\{\mathcal{P},\emptyset\}$ and one logical operator in $\{\mathcal{C},\mathcal{D},\mathcal{N},\emptyset\}$ as one step of operations, where the $\emptyset$ means direct data flow without any operation. As an outcome, our model can handle arbitrary types of FOL queries, including triples~\citep{BetaE}, triples with auxiliary information~\citep{StarQE}, multivariate facts, etc., with great flexibility and extensibility.

\begin{figure*}[t]
\centering
\includegraphics[width=17cm]{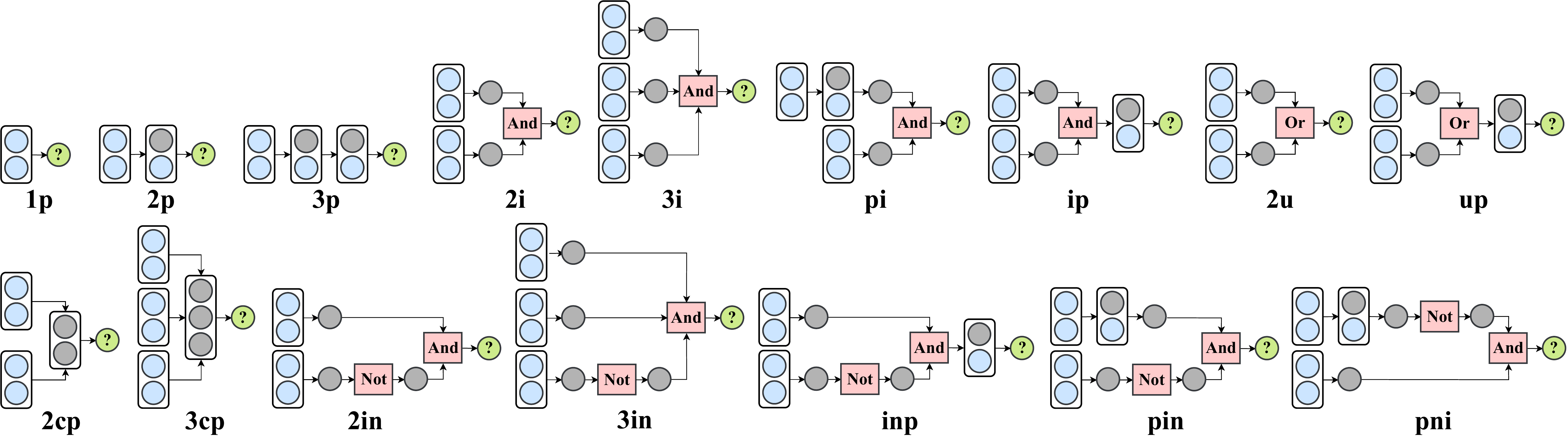}
\caption{There are 16 n-ary FOL queries in the WD50K-NFOL dataset, including 11 EPFO queries and 5 queries with negation. 2cp and 3cp are unique to our proposed n-ary fact query.}
\label{f4}
\end{figure*}

\begin{table*}[h!t]
\footnotesize
\setlength{\tabcolsep}{0.95mm}{
\begin{tabular}{ccccccccccccccccccc}
\toprule
\multicolumn{1}{c|}{\textbf{Model}}     & \textbf{AVG}$_p$        &\multicolumn{1}{c|}{ \textbf{AVG}$_n$}                   & \textbf{1p}          & \textbf{2p}          & \textbf{3p}          & \textbf{2i}          & \textbf{3i}          & \textbf{pi}          & \textbf{ip}          & \textbf{2u}          & \textbf{up}          & \textbf{2cp}         & \multicolumn{1}{c|}{\textbf{3cp}} & \textbf{2in}         & \textbf{3in}         & \textbf{inp}         & \textbf{pin}         & \textbf{pni}         \\ \midrule[0.8pt]
\multicolumn{19}{c}{\textbf{WD50K-QE}}                                                                                                                                                                                                                                                                                                                                                                                                                                                        \\ \midrule
\multicolumn{1}{c|}{StarQE-1p} & 26.04       & \multicolumn{1}{c|}{-} & 26.83       & 6.79                 & 19.72                & 56.16                & 74.35                & 39.81                & 9.62                 & -                    & -                    & -                    & \multicolumn{1}{c|}{-}            & -                    & -                    & -                    & -                    & -                    \\
\multicolumn{1}{c|}{StarQE}             & 64.47                & \multicolumn{1}{c|}{-}          & 30.98                & 44.13                & 52.96                & 63.14                & 83.78                & 71.52                & 55.20                & -                    & -                    & -                    & \multicolumn{1}{c|}{-}            & -                    & -                    & -                    & -                    & -                    \\
\multicolumn{1}{c|}{NQE-1p}             & 72.94                & \multicolumn{1}{c|}{-}          & \textbf{48.35}       & 48.25                & 59.79                & 70.90
& 85.55                & 77.63                & \textbf{60.79}       & -                    & -                    & -                    & \multicolumn{1}{c|}{-}            & -                    & -                    & -                    & -                    & -                    \\
\multicolumn{1}{c|}{NQE}                & \textbf{75.84}       & \multicolumn{1}{c|}{-}          & 47.69                & \textbf{50.31}       & \textbf{61.92}       & \textbf{76.86}       & \textbf{94.06}       & \textbf{81.32}       & 60.16                & -                    & -                    & -                    & \multicolumn{1}{c|}{-}            & -                    & -                    & -                    & -                    & -                    \\ \midrule[0.8pt]
\multicolumn{19}{c}{\textbf{WD50K-NFOL}}                                                                                                                                                                                                                                                                                                                                                                                                                                                               \\ \midrule
\multicolumn{1}{c|}{NodeH-only}   & 31.32          & \multicolumn{1}{c|}{12.10}          & 46.72          & 31.87          & 32.55          & 46.30          & 77.82          & 59.70          & 29.49          & 14.93          & 24.38          & 33.68          & \multicolumn{1}{c|}{43.09}          & 7.78           & 42.98          & 20.09          & 7.79          & 8.64          \\
\multicolumn{1}{c|}{EdgeH-only}   & 29.05          & \multicolumn{1}{c|}{12.21}          & 41.96          & 30.60          & 30.61          & 37.76          & 63.77          & 47.83          & 27.14          & 16.03          & 24.59          & 32.75          & \multicolumn{1}{c|}{42.68}          & 8.36           & 39.05          & 20.71          & 7.73          & 8.44          \\
\multicolumn{1}{c|}{Logic-blind}  & 32.92          & \multicolumn{1}{c|}{7.75}           & 49.92          & 34.09          & 35.26          & 51.03          & 90.72          & 65.55          & 30.72          & 13.66          & 22.95          & 37.15          & \multicolumn{1}{c|}{48.18}          & 3.54           & 40.25          & 14.31          & 2.95          & 5.74          \\
\multicolumn{1}{c|}{Unparalleled} & 33.47          & \multicolumn{1}{c|}{12.45}          & 54.81          & 34.35          & 34.47          & 49.22          & 84.56          & 63.05          & 31.98          & 14.76          & 26.14          & 38.73          & \multicolumn{1}{c|}{49.51}          & 7.79           & 43.89          & 20.02          & 8.22          & 7.65          \\
\multicolumn{1}{c|}{NQE-1p}       & 32.80          & \multicolumn{1}{c|}{12.56}          & 53.75          & 33.52          & 34.97          & 46.96          & 80.22          & 60.86          & 33.89          & 13.95          & 25.13          & 37.21          & \multicolumn{1}{c|}{50.80}          & 9.10           & 45.99          & 19.45          & 7.86          & 7.91          \\
\multicolumn{1}{c|}{NQE}          & \textbf{36.87} & \multicolumn{1}{c|}{\textbf{14.06}} & \textbf{55.90} & \textbf{37.04} & \textbf{36.86} & \textbf{53.10} & \textbf{95.90} & \textbf{68.15} & \textbf{34.72} & \textbf{18.66} & \textbf{27.85} & \textbf{40.08} & \multicolumn{1}{c|}{\textbf{51.26}} & \textbf{10.92} & \textbf{48.67} & \textbf{22.14} & \textbf{8.91} & \textbf{9.67}\\
 
\bottomrule 
\end{tabular}}

\caption{The MRR results (\%) of answering the n-ary FOL queries. AVG$_p$ is the average MRR for EPFO queries, and AVG$_n$ is the average MRR for queries with negation. Results for StarQE  are taken from its original paper~\citep{StarQE}.}
\label{main}
\end{table*}

\section{Experiments}
In this section, we design sufficient experiments to explore whether NQE is consistent with the following research questions (RQs).
\textbf{RQ1:} Whether NQE is a state-of-the-art model of CQA over HKGs?
\textbf{RQ2:} Can NQE handle all n-ary FOL queries on HKGs flexibly?
\textbf{RQ3:} How does each component proposed by NQE work?
\textbf{RQ4:} Does NQE have good generalization ability when training only one-hop queries (1p)?
\textbf{RQ5:} Does NQE have good interpretability in real examples?
\subsection{Experimental Setup}
\subsubsection{Datasets.}
The only existing dataset for CQA tasks over hyper-relational knowledge graphs (HKGs) is WD50K-QE proposed by ~\citep{StarQE}, which is constructed from the HKG standard dataset WD50K~\citep{StarE}. However, it only adds auxiliary information to the CQA for triples and has seven basic conjunctive queries $(1p,2p,3p,2i,3i,pi,ip)$. Therefore, we propose the WD50K-NFOL dataset, also extracted from WD50K, to satisfy all n-ary FOL queries, containing 11 EFPO queries~\citep{Q2B} $(1p,2p,3p,2i,3i,pi,ip,2u,2p,2cp,3cp)$ and 5 logical queries with negation $(2in,3in,inp,pin,pni)$ are shown in Figure~\ref{f4}. In particular, 2cp and 3cp are n-ary FOL queries that can only appear when there are more than three entities in the query facts.

\subsubsection{Evaluation Protocol.}
In all experiments, we use the similarity function proposed in section~\ref{4.4} to compute similarity probabilities $S(q)$ between the target query $q$ and all entities and rank them from largest to smallest, then we follow~\citet{TransE} to filter out other correct answers to obtain the rank $\operatorname{rank} \left(v_{i}\right)$ for entities $v$. We choose mean reciprocal rank (MRR$(q)=\frac{1}{\|v\|} \sum_{v_{i} \in v} \frac{1}{\operatorname{rank}\left(v_{i}\right)}$) and Hits at K (H@$K(q)=\frac{1}{\|v\|} \sum_{v_{i} \in v} f\left(\operatorname {rank}\left(v_{i}\right)\right)$) evaluation indicators introduced by~\citet{TransE} to evaluate these rankings, with larger values representing higher rankings and better evaluation results.
\subsubsection{Baselines.}
Currently, StarQE~\citep{StarQE} is the only model that can handle facts with more than two entities in them. Therefore we compare it with the NQE model as an essential baseline. However, NQE can handle more n-ary FOL queries, so we set some ablation models to compare with NQE. NodeH-only and EdgeH-only denote the variants where the dual-heterogeneous Transformer retains only node heterogeneity or edge heterogeneity, respectively; logic-blind denotes the variant where the logical operators are unified into averaging operations. Unparalleled denotes the variant where no parallel processing algorithm is used. Moreover, StarQE-1p and NQE-1p represent that StarQE and NQE train only one-hop queries, respectively.

\subsection{Main Results (RQ1)} 
To answer RQ1, we conduct experiments and evaluations on WD50K-QE. Table~\ref{main} shows the predictions of the NQE model on WD50K for seven kinds of n-ary FOL queries. Previous SOTA model StarQE can only solve the CQA problem with only two atomic operations, n-ary projection (p) and conjunction (i), are included, and only the 'object' position is predicted in the n-ary projection, which is a part of the n-ary FOL queries. The experimental results show that NQE is surprisingly and substantially better than the existing optimal model StarQE for all seven queries. Among them, the 1p,2i, and 3i improvements are 16.71\%, 13.72\%, and 10.28\%, which are all more than 10\%, and the average metric improvement is 11.37\%, which is because the NQE dual-heterogeneous Transformer encoder learns the heterogeneous sequence structure of n-ary facts sufficiently and representatively. The fuzzy vector logic conforms to all logical axioms, which is not available with StarQE based on the DAG method. Therefore, we conclude that NQE is a state-of-the-art model for handling CQA problems over HKGs.

\subsection{CQA for All N-ary FOL Queries (RQ2)}
We evaluate the effectiveness of NQE in handling all kinds of n-ary FOL queries on the new dataset WD50K-NFOL. WD50K-NFOL contains rich n-ary FOL queries, and its n-ary projection has queries not only for objects but also for subjects and the value of any attribute in the qualifiers. Also, in terms of logical query types, it has the three most common logical categories: conjunction, disjunction, and negation. 
Table~\ref{main} shows the MRR evaluation results of NQE for all n-ary FOL queries, which perform well with the best three queries 3i, pi, and 1p reaching 95.90\%, 68.12\%, and 55.90\%, respectively. We derived the average metrics for 11 EFPO queries and 5 queries with negation separately and found that the queries with negation are more complex than EFPO. It can be seen that because NQE defines four atomic operators, which makes it flexible for any combination, it is the first CQA model that can handle n-ary, in fact, all logical queries, and can arbitrarily combine four atomic operators to get any real-world n-ary FOL queries with good flexibility, which answers RQ2.

\subsection{Ablation Study (RQ3)} 
To explore the effectiveness of the n-ary projection, the three fuzzy vector logic operators, and the batch algorithm within the NQE model, we designed several variants of NQE for comparison to answer RQ3. First, NodeH-only and EdgeH-only represent the dual-heterogeneous Transformer encoder with only one heterogeneity, and EdgeH-only has similar results to NodeH-only. Nevertheless, neither model is as good as the complete model of NQE, demonstrating that both heterogeneities are essential for acquiring n-ary facts. We then consider a variant of the fuzzy vector logic operator, Logic-blind, and using an aggregation function like average, the MRR of AVG$_p$ is 3.95\% lower than that of NQE and even worse in the query of AVG$_n$, 6.31\% lower, indicating the importance of logic operators that conform to the logic axioms. Finally, we compare Unparalleled and NQE with a non-parallel processing training method variant and determine that the parallel processing algorithm proposed in section~\ref{4.4} can effectively improve 3.40\% in AVG$_p$ and 1.61\% in AVG$_n$. Since NQE can combine these queries without distinguishing their learning category, entities belonging to different categories can be included within the same batch while achieving better stochastic learning for the fuzzy vector embedding. In addition, this batch of algorithms can be easily extended to handle more queries, not just those given in Figure~\ref{f1}, with good extensibility.

\subsection{Generalization (RQ4)} 
We trained only one-hop queries and tested all n-ary FOL queries to test the model's generalization ability. We denote the variant with only 1p trained in NQE and StarQE as NQE-1p and StarQE-1p, respectively. NQE-1p has a notable 47.09\% improvement in the average MRR over StarQE-1p, and NQE-1p is more effective than StarQE trained with all queries, with an 8.66\% improvement in the average MRR. Our flexible logic operator has improved the performance, which breaks the pattern of StarQE requiring message-passing learning for each computational graph. The logic operator is more flexible, has no learnable parameters, and can arbitrarily expand the type and order of logical operations by using operators that conform to more logical axioms, providing excellent generalization capabilities, which answers RQ4. Moreover, we found that on two datasets, our NQE-1p and NQE full models differed on MRR average by 2.71\% and 3.93\%, respectively, indicating that NQE-1p already has a strong characterization capability and training the entire n-ary FOL queries can improve NQE performance further.

\subsection{Case Study (RQ5)} 
\begin{figure}[t]
\centering
\includegraphics[width=8.6cm]{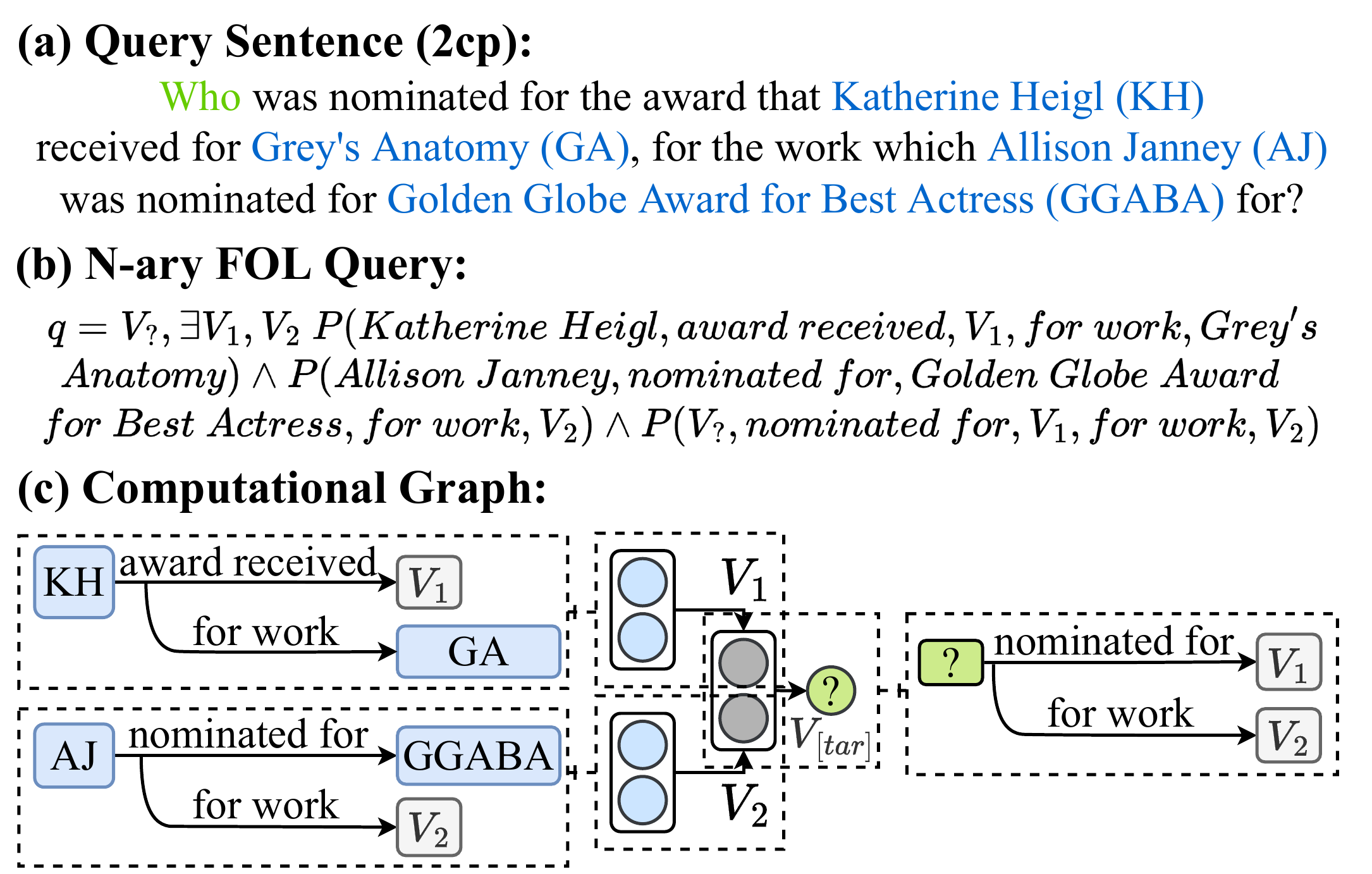}
\caption{Case study of `2cp' query type.}
\label{f6}
\end{figure}

To answer RQ5, we can use the similarity function to determine the entity prediction probabilities of the intermediate queries $V_1, V_2,\ldots, V_k$ and the target query $V_{[tar]}$ to make the whole prediction process interpretable. We present a case study with 2cp as an example as shown in Figure~\ref{f6}. In order to rank similar entities from largest to smallest, we compute the similarity probabilities of two intermediate queries and the target query sequentially in topological order from the computational graph based on the n-ary FOL query. We choose several answers to display and show their probabilities, correctness, and whether they are easy answers (answers in the test set), hard answers (answers in the test set), or wrong answers as shown in Table~\ref{example}. In this way, not only the answer to the final target query is obtained, but each intermediate inference step is also traceable.

\begin{table}[t]
\footnotesize
\centering

\setlength{\tabcolsep}{0.09cm}{
\begin{tabular}{cccccc}
\midrule[0.8pt]
\textbf{Var.}              & \textbf{Rank} & \textbf{Query Answers} & \multicolumn{1}{c}{\textbf{Pro.}} & \textbf{Cor.} & \textbf{Ans.} \\\midrule
\multirow{3}{*}{$V_1$} & 1       & Primetime Emmy Award..           & 78.35\%                                         & \ding{52}          & Easy        \\
                    & 2       & Golden Globe Award..            & 1.39\%                                          & \ding{56}          & -        \\
                    & 3       & Screen Actors Guild Award..            & 1.15\%                                          & \ding{52}          & Easy        \\\midrule
\multirow{3}{*}{$V_2$} & 1       & The West Wing             & 67.46\%                                          & \ding{52}          & Easy        \\
                    & 2       & Edie Falco             & 3.07\%                                          & \ding{56}          & -        \\
                    & 3       & My So-Called Life             & 1.75\%                                          & \ding{56}          & -        \\\midrule
\multirow{5}{*}{$V_{[tar]}$} & 1       & Allison Janney             & 42.38\%                                          & \ding{52}          & Easy        \\
                    & 2       & Stockard Channing             & 12.24\%                                          & \ding{52}          & Hard        \\
                    & 3       & Mary Steenburgen             & 7.02\%                                          & \ding{56}          & -        \\
                    & 4       & Janel Moloney             & 5.32\%                                          & \ding{52}          & Hard        \\
                    & 5       & Downton Abbey             & 1.97\%                                          & \ding{56}          & -    \\\bottomrule[0.8pt]
\end{tabular}}
\caption{Results of case study.}
\label{example}
\end{table}

\section{Conclusion}
In this paper, we propose a novel N-ary Query Embedding model, NQE, to handle all n-ary FOL queries for complex query answering over hyper-relational knowledge graphs. NQE uses a dual heterogeneous Transformer to handle n-ary projection and designs three logic operators using fuzzy vector logic theory, which has good flexibility and extensibility. The experiments show that NQE is a state-of-the-art model of the current CQA method for HKGs, with substantial progress in terms of metrics, the range of problems handled, and good generalization capability.

\section*{Acknowledgements}
This work is supported by the National Science Foundation of China (Grant No. 62176026, Grant No. 61902034) and Beijing Natural Science Foundation (M22009, L191012). This work is also supported by the BUPT Postgraduate Innovation and Entrepreneurship Project led by Haoran Luo.

\bibliography{aaai23}

\end{document}